\documentclass[letterpaper, 10 pt, journal, twoside]{ieeetran} 
\usepackage{amsmath,amsfonts}
\usepackage{algorithmic}
\usepackage{algorithm}
\usepackage{array}
\usepackage[caption=false,font=normalsize,labelfont=sf,textfont=sf]{subfig}
\usepackage{textcomp}
\usepackage{stfloats}
\usepackage{url}
\usepackage{verbatim}
\usepackage{graphicx}
\usepackage{cite}
\usepackage{times}
\usepackage{graphicx}
\usepackage{subcaption}  
\usepackage{multicol}
\usepackage{colortbl}
\usepackage{booktabs}
\usepackage{comment}
\usepackage{times}
\usepackage{epsfig}
\usepackage{lipsum}
\usepackage{amsmath}
\usepackage{pifont}
\usepackage{amssymb}
\usepackage{arydshln}
\usepackage{color}
\usepackage{url}
\usepackage{here}
\usepackage{balance}
\usepackage{xspace}
\usepackage{multirow}
\usepackage{bbding}
\usepackage{stfloats}
\usepackage{wrapfig}
\usepackage{placeins}
\usepackage{hyperref}

\newcommand{\MethodName}{MBRA}

\newcommand{\DatasetName}{FrodoBots-2k}

\newcommand{\PolicyName}{LogoNav}

\begin{document}

\title{Learning to Drive Anywhere with Model-Based Reannotation}

\author{Noriaki Hirose$^{1,2}$, Lydia Ignatova$^{1}$, Kyle Stachowicz$^{1}$, Catherine Glossop$^{1}$, Sergey Levine$^{1}$ and Dhruv Shah$^{1,3}$%

\thanks{$^{1}$N. Hirose, L. Ignatova, K.Stachowics, C. Glossop, S. Levine and D. Shah are with Department of Electrical Engineering and Computer Sciences , University of California, Berkeley, CA USA
        {\tt\footnotesize noriaki.hirose@berkeley.edu}}%
\thanks{$^{2} $N. Hirose is with Toyota Motor North America, Inc, Ann Arbor MI USA}
\thanks{$^{3} $D. Shah is with Department of Electrical and Computer Engineering, Princeton University, Princeton, NJ USA}
}

\makeatletter
\let\@oldmaketitle\@maketitle%
\renewcommand{\@maketitle}{\@oldmaketitle%
    \centering
    \vspace{6mm}
    \includegraphics[width=0.99\linewidth]{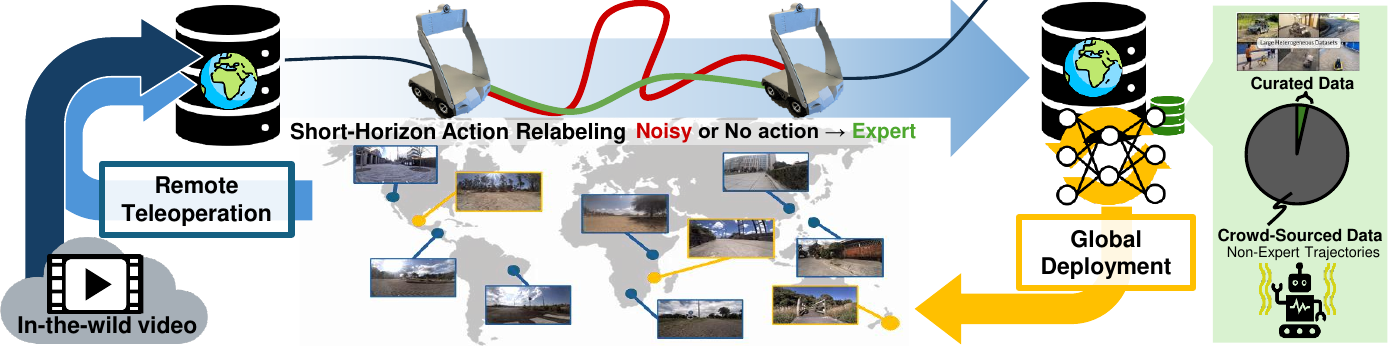}
    \captionof{figure}{We train a highly generalizable navigation policy that can control robots in a variety of conditions and be deployed zero-shot in new environments across the world. Our proposed method, \textbf{M}odel-\textbf{B}ased \textbf{R}e\textbf{A}nnotation, enables imitation learning from noisy, passive data, such as low-quality crowd-sourced demonstrations or even videos from the web.}
    \label{fig:overview}
}
\addtocounter{figure}{-2}



\maketitle

\begin{abstract}
Developing broadly generalizable visual navigation policies for robots is a significant challenge, primarily constrained by the availability of large-scale, diverse training data. While curated datasets collected by researchers offer high quality, their limited size restricts policy generalization. To overcome this, we explore leveraging abundant, passively collected data sources, including large volumes of crowd-sourced teleoperation data and unlabeled YouTube videos, despite their potential for lower quality or missing action labels. We propose Model-Based ReAnnotation (MBRA), a framework that utilizes a learned short-horizon, model-based expert model to relabel or generate high-quality actions for these passive datasets. This relabeled data is then distilled into LogoNav, a long-horizon navigation policy conditioned on visual goals or GPS waypoints. We demonstrate that LogoNav, trained using MBRA-processed data, achieves state-of-the-art performance, enabling robust navigation over distances exceeding 300 meters in previously unseen indoor and outdoor environments. Our extensive real-world evaluations, conducted across a fleet of robots (including quadrupeds) in six cities on three continents, validate the policy's ability to generalize and navigate effectively even amidst pedestrians in crowded settings. We present videos showcasing the performance and release our checkpoints and training code on our project page\footnote{\bf \href{https://model-base-reannotation.github.io/}{\texttt{model-base-reannotation.github.io}}}.
\end{abstract}


\section{Introduction}
\IEEEPARstart{M}{achine} learning has demonstrated remarkable success across a range of tasks, including natural language processing~\cite{vaswani2017attention,brown2020language} and computer vision~\cite{radford2021learning,kirillov2023segment,oquab2023dinov2}. A key factor driving these advancements is the availability of large and diverse training datasets. In robotics, lack of data is a major bottleneck: intentional, centralized data-collection efforts are costly, requiring real-world robots and human operators, while Internet-scraped data is rarely directly applicable to the robotics domain~\cite{o2023open,khazatsky2024droid}.

In this paper, we study the problem of developing an end-to-end robot navigation policy capable of generalizing to a wide range of outdoor and indoor environments and navigating to distant goals hundreds of meters away.
Training such an end-to-end policy requires large amounts of diverse data to grant broad coverage over possible environments. Previous navigation works~\cite{shah2023gnm} have relied on centrally collected datasets generated by robotics researchers. While these datasets tend to be high quality, the sum total of these datasets is on the order of dozens of hours~\cite{shah2023vint}, limiting the breadth of generalization that can be achieved from this high-quality data alone.

Facing this data limitation, we turn our attention to making use of more abundant sources of \textit{passive data} -- data that lacks actions or only provides low-quality action labels. For example, crowd-sourced data, collected in a decentralized fashion by a large user base, has high state coverage and a diverse set of environments compared to what can be collected in a centralized fashion. However, the challenging nature of remote data collection with non-expert demonstrators makes it difficult to train good policies directly on the actions in such datasets. In-the-wild video is another passive data source that contains diverse environments and can enable more generalized performance. However, in-the-wild video does not have associated actions at all. 

To enable the use of these cheap, scalable data sources, we propose robust model-based learning to train a short-horizon expert \textit{relabeling model} for generating high-quality actions connecting two nearby states. We use this short-horizon relabeling model to annotate actions in the passive dataset, which then gives us much cleaner and higher-quality actions than in the original dataset.
The outputs of this relabeling model are then distilled into the long-horizon policy that can be conditioned on visual goals or on a future GPS waypoint for navigating over long distances. 

We deploy our system in a comprehensive set of evaluations across a fleet of low-cost robots deployed globally as well as various embodiments including the quadruped robot and find that it is able to deliver strong generalized performance in six different cities across three continents.

Our primary contributions are 1) a framework to learn a well-generalized long-horizon policy by applying a short-horizon relabeling \MethodName{} (\textbf{M}odel-\textbf{B}ased \textbf{R}e\textbf{A}nnotation) model to the passive data,
2) an instantiation of the \MethodName{} relabeler on the \DatasetName{} dataset and YouTube videos, yielding a strong short-horizon policy that we evaluate in 6 countries, and
3) \PolicyName{} ({\bf Lo}ng-range {\bf Go}al Pose-conditioned {\bf Nav}igation policy), a policy trained with \MethodName{} that achieves robust goal-reaching capabilities at 300+ meter scales, even while navigating around pedestrians in crowded environments. Please see our supplemental materials for videos of \PolicyName{} exhibiting robust driving behavior in complex long-horizon navigation settings.

%
\begin{figure*}[t]
\centering
\includegraphics[width=0.99\hsize]{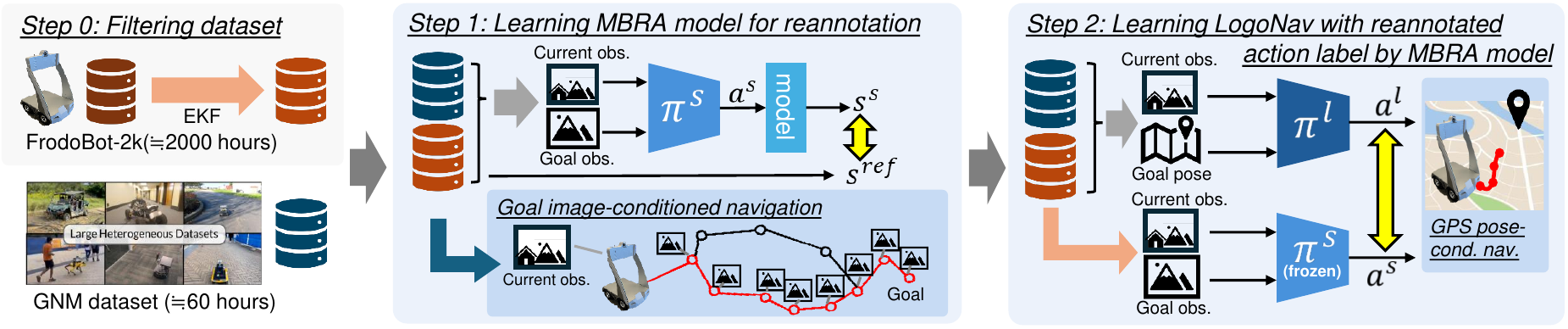}
\caption{{\small\textbf{Overview of \MethodName.} We propose a two-step process: In the first stage, we train a short-horizon reannotation policy with a robust MBL approach on the noisy dataset, which can be used for short-horizon image-conditioned navigation and which we leverage to relabel the noisy dataset with improved action labels. In step 2, we train a long-horizon navigation policy with the generated action labels.}}
\label{f:system_overview}
\end{figure*} 
\section{Related work}
\noindent\textbf{Vision-based robot navigation} has been widely explored to navigate toward goal positions given visual observations from a monocular camera. \cite{savinov2018semi,pathak2018zero,hirose2019deep} train short-horizon policies to generate actions with access to a single goal observation. These short-horizon policies often utilize topological memory to extend the range of navigation \cite{meng2020largescaletopo}. Some works\cite{shah2021rapid} use exploration with a topological memory to seek out a distant image goal, while others\cite{shah2022viking,stachowicz2023fastrlap} use a GPS signal for localization and navigate toward a goal provided as a 2D position in cartesian coordinates. Goal images and poses require prior access to the target environment and knowledge of the environment's geometry. 
Various learning methodologies such as imitation learning (IL)~\cite{savinov2018semi,shah2023gnm,shah2023vint}, reinforcement learning (RL)~\cite{stachowicz2023fastrlap,hiroseselfi,stachowicz2024lifelong}, and model-based learning (MBL)~\cite{hirose2019deep,hirose2023exaug} have been explored for training goal-conditioned vision-based policies on publicly available robot datasets.

These methods require a sequence of image observations and corresponding actions parsed from accurate wheel odometry~\cite{hirose2019deep,hirose2018gonet}, GPS~\cite{shah2021rapid}, and other reliable sensors. These datasets are collected via intentional, centralized teleoperation efforts with the downstream goal of training a navigation policy and, therefore, contain goal-directed trajectories. Collecting data of this sort at a global scale would require a massive unified effort that would be costly and time-consuming. 

\noindent\textbf{Robot learning with passive data.}
Visual SLAM~\cite{liu2024citywalker} and inverse dynamics models (VPT)~\cite{baker2022videopretrainingvptlearning} can be used to estimate trajectories for first-person videos, allowing us to train policies that use these trajectories as approximations of robot actions from action-free and non-robot data. While visual SLAM and its successors~\cite{orbslam3, mast3rslam, droidslam, dpvo} offer good local trajectory estimation, its accuracy relies on having consistent, good visual features in the image view.

Prior methods have also sought to leverage suboptimal data, including data without actions, by training a separate model to infer the action given the current and next state~\cite{baker2022videopretrainingvptlearning}. We find that this approach performs worse than MBRA on the highly suboptimal datasets we consider, both because the action prediction must leverage other datasets that contain distributional shift. Perhaps even more importantly, MBRA generates synthetic actions that are optimized to reach future states in a trajectory while satisfying basic navigational objectives via a forward model, rather than greedily maximizing the probability of the action given a pair of adjacent states, leading to smoother and more reasonable behavior.

\begin{figure}[t]
\centering
\includegraphics[width=1.0\hsize]{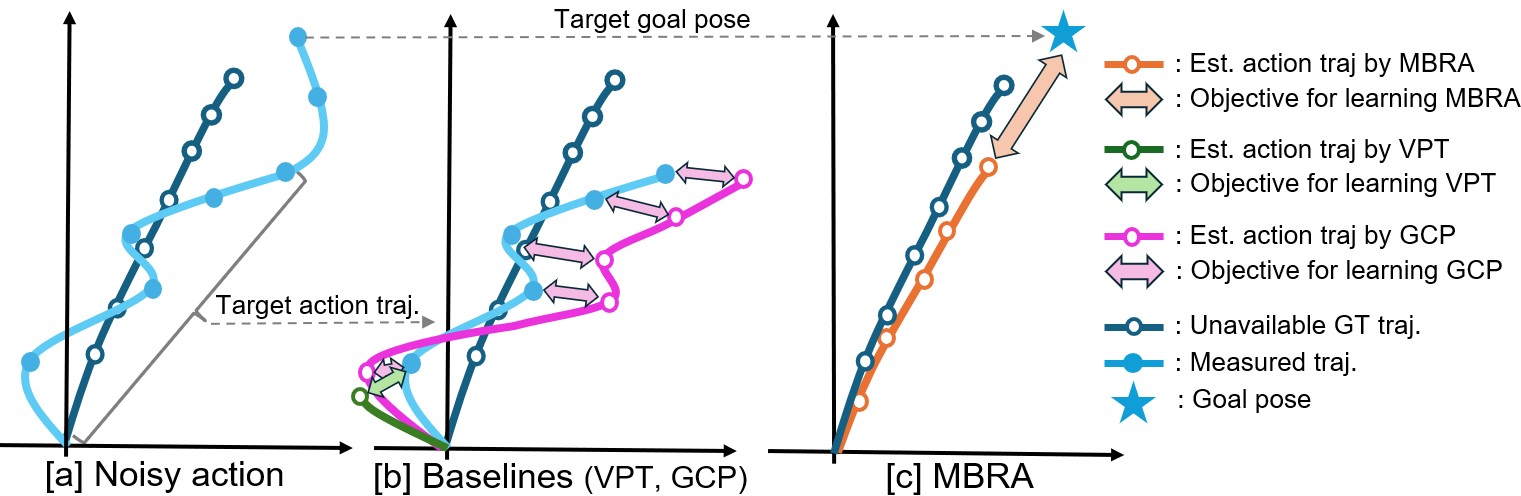}
\caption{{\small The aim of MBRA is to relabel low-quality data with actions that are \emph{better} than the actions in the dataset, in the sense that they more effectively link states over short-horizon trajectory snippets. Compared to methods that use one- or multi- step inverse models (e.g., VPT, GCP) or the original noisy actions, training on actions from MBRA leads to significantly more effective policies.}}
\label{f:annotation}
\end{figure}

Related to inverse model methods such as VPT, multi-step goal-conditioned policies~\cite{shah2023vint,sridhar2024nomad} (GCPs) train a model to predict the action given the current state and a (more distant) future state. While these approaches can also take non-greedy actions, unlike MBRA, they do not optimize for actions that satisfy navigational objectives. They also still suffer from distributional shift when the action labels in the target domain are unavailable, or else must use noisy action labels if they are present, both of which degrade performance (see Fig.~\ref{f:annotation}[b]). In our experiments, we find that MBRA significantly outperforms such methods when using large datasets with low-quality actions.

\section{Learning Short-Horizon Relabeling Policies with Model-based Learning}
\label{sec:reannotation}
In this paper, we focus on learning a long-horizon navigation policy from a highly diverse but suboptimal dataset $\mathcal{D}_{n}$. In particular, we wish to learn high-quality navigation from a crowd-sourced dataset; this requires us to train a relabeler that can predict actions that are better than those found in the original dataset. We assume access to a smaller \textit{clean} dataset $\mathcal{D}^*$ that contains high-quality behavior, such that $\lvert\mathcal{D}^*\rvert<\!\!<\lvert\mathcal{D}_n\rvert$.

While observations in $\mathcal{D}_n$ might represent high state coverage, the actions are low-quality: both because of inaccuracy due to state estimation errors and heterogeneity of uncurated human operators with varying skill levels. Our key insight is that a short-horizon model-based \emph{expert} trained from these noisy datasets can be used to relabel entire trajectories. This leads to a high-quality training dataset for an end-to-end navigation policy that can imitate these clean actions (Fig.~\ref{f:system_overview}).
\subsection{Learning a Short-Horizon Relabeling Model, \MethodName{}}
To train an accurate \MethodName{} model $\{ a^{s}_i \}_{i=0 \ldots N-1} = \pi^s(O_c; O_g)$ to infer the optimal actions occurring between the current observation $O_c$ and the goal observation $O_g$ for re-annotation, we propose the robust MBL to leverage the entire suboptimal dataset $\mathcal{D}_n$ as well as $\mathcal{D}^*$ in training. 

\noindent\textbf{Overview of learning MBRA model.}
As illustrated in Fig.~\ref{f:annotation}[b], single-step VPT and multi-step GCPs struggle to learn from noisy action labels during training.
To address this, we use Model-based Learning (MBL). MBL prioritizes the final goal state, but rather than mimicking the potentially noisy actions in the dataset, it utilizes a forward model to generate synthetic actions. Crucially, these actions are optimized both to reach future states in the trajectory and to satisfy basic navigational constraints (as shown in Fig.~\ref{f:annotation}[c]). This approach leads to smoother and more reasonable behavior, allowing MBL to leverage the noisy \DatasetName{} more effectively and resulting in better labels for the entire \DatasetName{} dataset. This allows us to improve the action label quality, while also preserving teleoperators' intent, such as avoiding collisions with pedestrians, staying on paths, etc.

\noindent\textbf{Learning architecture based on MBL.}
We design the following model-based objective $J_{mbl}$ for learning $\pi^s$ to reach target state $s^{ref}$ with keeping basic navigation constraints such as avoiding collision. Note that we only give the further target state $s^{ref}$ instead of giving the individual adjacent target states such as $s^{ref}_i$ in each $i$-th step not to be sensitive for low-quality states in the dataset as shown in Fig.~\ref{f:annotation}[b]. 
\begin{equation}
    \min \,J_{mbl} := \sum_{i=0}^{N-1}(s^{ref} - s^{s}_i)^2,
    \label{eq:MBL_objective}
\end{equation}
where $\{ s^{s}_i \}_{i = 0 \ldots N-1}$ are the estimated states at each step. The states $\{ s^{s}_i \}_{i = 0 \ldots N-1}$ are calculated by computing rollouts through a differentiable forward dynamic model $f$. The forward model considers the current observation $O_c$ and generated actions $\{ a^{s}_i \}_{i=0 \ldots N-1}$ from the short-horizon \MethodName{} model, $\pi^s$~\cite{hirose2022depth360}.
\begin{equation}
     \{ s^{s}_i \}_{i = 0 \ldots N-1} = f (O_c, \{ a^{s}_i \}_{i=0 \ldots N-1}).
     \label{eq:state estimation}
\end{equation}

While the states $\{ s^{s}_i \}_{i = 0 \ldots N-1}$ are conditioned on actions $\{ a^{s}_i \}_{i=0 \ldots N-1}$ and $f$ is differentiable, we can calculate the gradient of $\pi_s$ to minimize $J_{mbl}$ in each training step and learn $\pi_s$ by repetitively update the parameters of $\pi_s$ similar to other machine learning approaches. We do not modify the forward dynamics model $f$ during this training $\pi_s$. Detail implementation is shown in the Sec.\ref{sec:shortnav}.

Note that while we are still training $\pi^s$ on the suboptimal dataset in addition to $\mathcal{D}^*$, we are not directly imitating actions. By relying on the forward model $f$ and a reasonable distant target state $s^{ref}$, we can mitigate the effects of both the suboptimal action labels as well as noisy tracking information (Fig.~\ref{f:annotation}[c]). Therefore, the \MethodName{} model can leverage the visually and behaviorally diverse dataset, despite the low-quality actions. 
\subsection{Learning a Long-Horizon Navigation Policy, \PolicyName}
To train our long-horizon navigation policy, we first re-annotate the crowd-sourced dataset $\mathcal{D}_n$ with our learned $\pi^s$. This gives us a clean set of action labels that can be distilled into an end-to-end navigation policy $\pi^l$. We want a navigation policy $\pi^{l}$ to predict actions as $\{ a^{l}_i \}_{i=0 \ldots N-1} = \pi^{l} (O_c, p_g)$ where $O_c$ is the current observation and $p_g$ is the 2D relative goal pose from the robot coordinate. Notably, $p_g$ is at least 10 times further than the usual goal pose for the short-horizon relabeling model, on the order of 50 meters, compared to the previous 3 meters. We train this policy using imitation learning on the re-annotated action commands $\{ a^{s}_i \}_{i=0 \ldots N-1}$ from the short-horizon relabeling model such as $\min \,J_{il} := \sum_{i=0}^{N-1}(a^{s}_i - a^{l}_i)^2$.
By imitating the cleaned action commands linking $O_c$ and $O_g$, our long-horizon policy, \PolicyName~, can learn navigational affordances, such as staying on paths, avoiding collisions, and not disturbing pedestrians, which is representative of the ``good'' navigation behavior modeled by the \MethodName{} model. 
Note that we co-train on the relabeled $\mathcal{D}_n$ as well as the high-quality dataset $\mathcal{D}^*$. We freeze $\pi^s$ while training $\pi^l$.

\section{Implementation}
\label{sec:implementation}
We provide the implementation details of our navigation system, covering the dataset used, network and objective design, and hyperparameter settings used for training and dataset preparation.

\subsection{Passive Dataset}

We evaluate our approach with two different passive datasets, a crowd-sourced robotic dataset, \DatasetName{}, and an in-the-wild YouTube video dataset described in \cite{hiroselelan}. We focus on results using \DatasetName{} to demonstrate the effectiveness of our proposed approach and additionally evaluate its capabilities on the YouTube video dataset. 
\begin{table}[t]
  \begin{center}
  \vspace{2mm}
  \caption{Survey of public datasets for learning vision-based navigation policies in real-world.}  
  \label{tab:data} 
  \resizebox{0.9\columnwidth}{!}{
  \begin{tabular}{lccl} \toprule 
    Dataset & Policy & hour & Sensors\\ \midrule
    KITTI odom~\cite{Geiger2012CVPR} & teleop & 0.7 & RGB, 3D LiDAR, GPS \\
    NCLT~\cite{carlevaris2016university} & teleop & 34.9 & RGB, 3D LiDAR, odom, GPS, IMU\\ 
    GO Stanford~\cite{hirose2018gonet,hirose2019deep} & teleop & 10.3 & RGBs, odom\\
    FLOBOT~\cite{yan2020robot} & auto & 0.46 & RGBD, 3D and 2D LiDAR, odom, IMU.\\
    RECON~\cite{shah2021rapid} & auto & 25.0 & stereo RGBD, 2D LiDAR, GPS, IMU\\
    JRDB~\cite{martin2021jrdb} & teleop & 1.1 & stereo RGBD, 3D and 2D LiDAR, IMU\\
    SCAND~\cite{karnan2022socially} & teleop & 8.7 & RGBD, 3D LiDAR, odom\\ 
    TartanDrive~\cite{triest2022tartandrive} & teleop & 5.0 & RGBD, GPS, IMU\\     
    HuRoN~\cite{hirose2023sacson} & teleop & 75.0 & RGBs, 2D LiDAR, odom, bumper\\ \midrule    
    \DatasetName{} & teleop & 2000 & RGBs, GPS, IMU, odom,\\
    \DatasetName{}-filtered & teleop & 700 & RGBs, filtered 2D localization\\\bottomrule
  \end{tabular}%
  }
  \end{center}
\end{table}

\noindent
\textit{{\bf Crowd-sourced robotic dataset: }} 
The \DatasetName{} dataset~\cite{frodobots_2K} includes 2000 hours of data from over 10 cities and was collected as part of FrodoBots AI, where users explore locations worldwide by teleoperating robots to reach target positions. The \DatasetName{} dataset is significantly larger than other publicly available datasets for vision-based navigation tasks.  
As shown in Table \ref{tab:data}, the full version of the \DatasetName{} is more than 25 times larger than other datasets and includes a diverse set of real robot trajectories teleoperated by humans.

While the scale and diversity of this dataset are enticing, the inexpensive hardware setup of the robots and crowd-sourcing approach result in significant noise. Since sensor measurements cannot be reliably used to estimate robot poses, policies trained on raw actions have poor performance. 
The main factors of noisy action labels are 1) robot inconsistencies and corresponding user adjustments, 2) low-cost GPS and IMU, 3) inevitable wheel slips during turning, 4) robot vibration during turning, and 5) system delay. Details of the robot system are shown in \cite{erz_sdk} and Sec.~\ref{sec:evaluation_setup}.

\noindent
\textit{{\bf In-the-wild YouTube videos: }}
We also evaluate the ability of \MethodName{} to enable the use of non-robot data. We reannotate 100 hours of action-free in-the-wild YouTube videos, listed in \cite{hiroselelan}, and train a version of \PolicyName{} with the generated actions. These videos include inside and outside walking tours from 32 different countries across varying weather conditions, time, and environment types (urban, rural, etc.).

In addition to the passive data, we use the public expert datasets RECON~\cite{shah2021rapid}, GO Stanford~\cite{hirose2018gonet,hirose2019deep}, CoryHall~\cite{kahn2018self}, TartanDrive~\cite{triest2022tartandrive}, HuRoN~\cite{hirose2023sacson}, Seattle~\cite{shaban2022semantic}, and SCAND~\cite{karnan2022socially} with accurate action labels. The weighting of each dataset is the same as the original GNM dataset mixture.
\subsection{Pre-Processing and Filtering}
As shown on the leftmost side of Fig.~\ref{f:system_overview}, we use a classical state estimation pipeline to get better coarse robot pose estimates for \DatasetName{}. We use a smoothing system based on a bidirectional Extended Kalman Filter (EKF)~\cite{kalman1960new} to fuse raw actions with wheel speed measurements, GPS location, and compass heading (all of which are noisy) to get a smoothed estimate of the robot's position. We also filter out data where the robot is paused for a long time to prioritize learning desirable behaviors. The cleaned and filtered data consists of approximately 700 hours of real-world navigation trajectories collected worldwide, which is still an order of magnitude larger than any currently available visual navigation dataset as shown in Table~\ref{tab:data}.
While the EKF-based state estimation helps produce a less noisy action estimate~\cite{kfsurvey}, the signal remains too noisy for direct training. 
\subsection{Training Details}
We describe the training settings for both our short-horizon relabeling model, and long-horizon navigation policy.

\noindent
\textit{{\bf Short-horizon relabeling model: }} 
Following \cite{hirose2023exaug}, to encourage the \MethodName{} model to smoothly connect between $O_c$ and $O_g$ without collision, we design $s^{s}_i$ by three components, $[\hat{p}_i, \hat{c}_i, \Delta a^s_i]$,  Here $\hat{p}_i$ is the $i$-th virtual robot pose, $\hat{c}_i$ is the estimated collision state at $i$-th virtual robot pose $\hat{p}_i$ (where zero indicates no collision), and $\Delta a^s_i$ indicates the action difference, $a^s_{i+1} - a^s_{i}$. Accordingly, we define $s^{ref}$ as [$p_g$, 0.0, 0.0], where $p_g$ is further goal pose. 
 
Since we design $\{ a^{s}_i \}_{i=0 \ldots N-1}$ as the linear and angular velocities, we calculate the unicycle model to integrate the velocity commands for $N$ = 8 steps at 3 Hz and generate the virtual robot poses $\{\hat{p}_i \}_{i=0 \ldots N-1}$ in the dynamic forward model $f$. In addition, we estimate 3D point clouds from the current image $O_c$ via the depth estimation model~\cite{hirose2022depth360} and count collided 3D points at $\{\hat{p}_i\}_{i=0 \ldots N-1}$ as $\{\hat{c}_i\}_{i=0 \ldots N-1}$ in $f$. By penalizing $\hat{c}_i$ to be smaller values, the MBRA model learns to generate collision-free actions. 
While our objectives are not explicitly designed to learn semantic behaviors such as staying on paths or avoiding pedestrians, the generated trajectories effectively connect $O_c$ and $O_g$, implicitly capturing such behaviors. Although the action labels are unusable due to heavy noise, the image sequences from teleoperation still reflect the teleoperator's semantic intent. 

Furthermore, since the robot system has $L$ steps system delay~\cite{cui2023learning,alnajdi2023machine} when operating remote robot via internet, we design our objective and network architecture to account for system delay to prevent overshooting or oscillating around target trajectories. 
Inspiring the previous works of model predictive control~\cite{kwon2004general,kwon2003simple}, we consider the robotic states with the previous action commands $\{ a_i \}_{i=-L \ldots -1}$ to genrate the actions $\{ a_i \}_{i=0 \ldots N-1}$. 

In training, we set the observation and action rate for trajectory sampling at 3 Hz for consistency with the GNM dataset. During training, we randomly select an image frame from the entire dataset as the current observation, and then randomly select a goal frame from up to $N_g =$ 20 steps (about 7 seconds) in the future. This short distance to the goal lets us learn precise labels to reannotate the action between $O_c$ and $O_g$. 
More details are shown in \cite{hirose2023exaug} and our supplemental code base. 

\noindent
\textit{{\bf Long-horizon navigation policy: }} 
For long-horizon navigation, we use a larger $N_g = 100$ to sample a goal position up to 33 seconds into the future. We reannotate actions with the short-horizon \MethodName{} model to get high-quality action labels for the \DatasetName{} dataset. This process yields action labels with a chunk size of $N=8$ steps. We train on the IL objective $J_{il}$ using the same parameters and settings as the short-horizon relabeling model otherwise. Since the action space for long-horizon navigation is the 2D pose, following \cite{shah2023vint}, we use the integrated pose commands from the \MethodName{} model as the supervision. In inference, we apply the PD controller to calculate the velocity commands from the generated target pose, similar to \cite{shah2023vint,sridhar2024nomad}. The observation space of $O_c$ and $O_g$ is the image space for both policies. 

\begin{figure}[t]
\centering
\includegraphics[width=0.99\hsize]{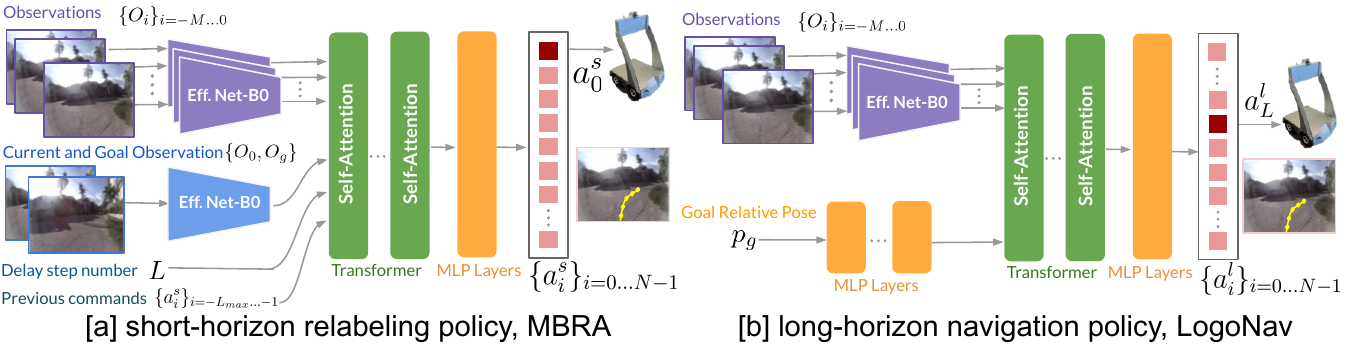}
\caption{{\small\textbf{Network architecture.} In addition to the visual observations, We feed the delay step and the previous actions to consider the system delay in the MBL objective. For the long-horizon navigation policy, we replace the visual encoder for the current and the goal observation with the MLP layers for the goal pose.}}
\label{f:network}
\end{figure}

\noindent
\textit{{\bf Network design: }} Figure \ref{f:network} shows the network architecture of both our \MethodName{} model, $\pi^{s}$ and \PolicyName{} policy, $\pi^{l}$.
For $\pi^{s}$, we concatenate the current observation $O_c$ and the goal observation $O_g$ and generate a goal-conditioned embedding with EfficientNet-B0. In addition, we concatenate the image observation history $\{O_i\}_{i = -M \ldots 0}$ and generate a history embedding with EfficientNet-B0. We pass in these visual features, the system delay $L$  and the previous action commands $\{ a^s_i \}_{i=-L \ldots -1}$ to a set of Transformer and fully connected MLP layers to produce action commands $\{a^s_i\}_{i=0 \dots N-1}$. 

For $\pi^{l}$, we replace the visual encoder for $O_c$ and $O_g$ with MLP layers for the local goal pose $p_g$ on the current robot coordinate in our implementation. In addition, we no longer include system delay length $L$ and previous actions $\{ a^s_i \}_{i=-L \ldots -1}$. Instead of considering the delay during training, we use the $L^{\textrm{th}}$ step of the output during inference, similar to \cite{chi2024universal} and \cite{shah2023vint}. More details are shown in the supplemental code base. 
\section{Evaluation}
To evaluate \PolicyName{} and the impact of \MethodName{} relabeling in the real world, we focus our experiments on answering the following questions:
\vspace{0mm}
\begin{itemize}
    \setlength{\parskip}{0cm}
    \setlength{\itemsep}{0cm} 
    \item[{\bf Q1}] Can we apply \MethodName{} to learn an effective long-horizon navigation policy?
    \item[{\bf Q2}] Can we use \MethodName{} for action-free in-the-wild data? 
    \item[{\bf Q3}] Is \MethodName{} more effective at learning relabeler from low-quality datasets than IL? 
\end{itemize}
\subsection{Evaluation Setup}
\label{sec:evaluation_setup}
We describe both the short-horizon and long-horizon navigation tasks, as well as the robot hardware, on which we evaluate our method, along with their associated baselines.

\noindent
\textit{{\bf Short-horizon navigation policy: }} 
Short-horizon navigation policy (\MethodName{} model) can navigate the robot toward a goal up to 3 meters away, so we use a topological memory to enable the robot to navigate to further goal positions, similar to other vision-based navigation approaches~\cite{savinov2018semi,hirose2019deep,shah2023vint}. To collect this goal loop, we teleoperate the robot and record image observations at a fixed frame rate of 1 Hz. To deploy the policy, we start from the initial observation and continuously estimate the closest node as the current node at each time step, following \cite{shah2023vint,sridhar2024nomad}. We feed the image from the next node as the goal image $O_g$ to our policy to compute the next action. 

\noindent
\textit{{\bf Long-horizon navigation policy: }} 
Our \PolicyName{} policy can navigate to goals between 25-100 meters from the initial robot pose in environments unseen during training. 
We rely on GPS(outdoor) and tracking camera~\cite{t265}(inside) to get robot positions and specify goals. We evaluate longer trajectories by setting multiple subgoals at intervals of approximately 80 meters apart. At every step, we calculate the relative goal pose $p_g$ on the way to the next goal pose. When $|p_g| < 5.0$ m, we consider the goal reached and update to the next subgoal for a longer trajectory. 

\begin{figure}[t]
\centering
\includegraphics[width=0.99\hsize]{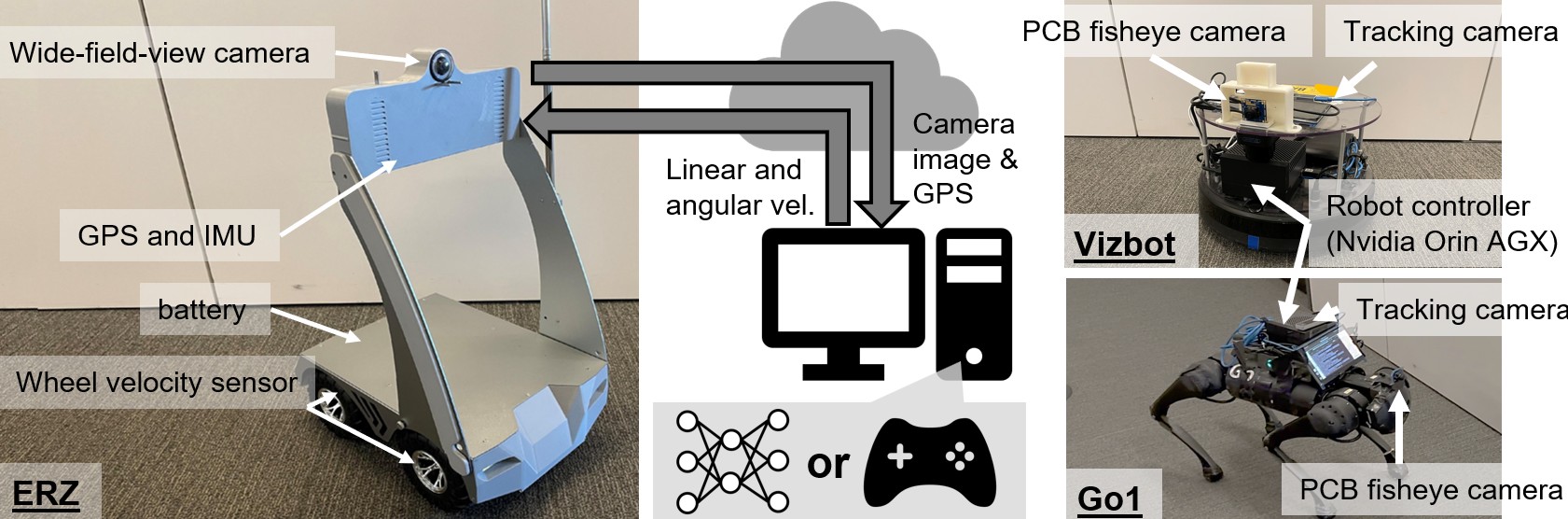}
\caption{{\small\textbf{Overview of the robot hardwares and systems.} ERZ can be controlled over a internet connection for data collection and for deploying our navigation policy. Vizbot and Go1 with different cameras are controlled from the onboard robot controller with ROS.}}
\label{f:robots}
\end{figure}

\noindent
\textit{{\bf Robot hardware: }} 
The FrodoBot ``Earth Rover Zero" (ERZ), shown in Fig.~\ref{f:robots}, is a low-cost RC car used both for the \DatasetName{} dataset collection and our main evaluation. The ERZ includes a host of sensors such as front and back side cameras, GPS, an IMU unit including gyroscope, accelerometer and compass sensors, and wheel velocity sensors in all four wheels. All measurements from the sensors can be accessed through the platform's API. Linear and angular velocity commands can also be sent to the robot from user teleoperation for data collection (gaming) or from our trained policies for navigation. The rover can turn in place and lasts for five hours on a fully charged battery.

In addition, we conduct additional evaluations with different robot hardware and systems with the wheeled mobile robot, VizBot~\cite{niwa2022spatio} and the quadruped robot, Unitree Go1 to analyze the cross-embodiment performance of our policy. 
\subsection{Baseline Method}
In our evaluation, we use the following two baselines, NoMaD and behavior clolning (BC). For BC, we evaluate various annotation methods as the ground truth action labels to compare with our \MethodName{} model.

\noindent
\textit{{\bf NoMaD~\cite{sridhar2024nomad}:}} We deploy the NoMaD policy \cite{sridhar2024nomad} for exploration and generate 30 possible trajectories. Out of these options, we select the best trajectory by measuring the distance between the last predicted position and the goal pose and selecting the minimum one to control the robot.

\noindent
\textit{{\bf Behavior Cloning~\cite{shah2023vint}:}} We train a long-horizon navigation policy on reannotated action labels by following several baseline methods instead of using our \MethodName{}. Similar to our methods, we sample 8 steps robot trajectory at 3.0 Hz in the current robot coordinates in all baseline methods. All learning setups except annotation are same as our method.

\setlength{\leftskip}{0.3cm}
\noindent
\textit{{\bf Raw action label:}} As the simplest action commands, we annotate the robot trajectory by integrating the teleoperator's velocity command. 

\noindent
\textit{{\bf Filtered action label:}} We give the mentioned EKF for entire \DatasetName{} dataset to estimate the less-noisy robot pose. We transform them into the local robot coordinate.

\noindent
\textit{{\bf Visual SLAM~\cite{liu2024citywalker}:}} Following \cite{liu2024citywalker}, we estimate the global trajectories with one of the state-of-the-art viusal SLAM, DPVO~\cite{dpvo}. To have better pose estimation, we rectify all images in \DatasetName{} dataset for DPVO. 

\noindent
\textit{{\bf VPT~\cite{baker2022videopretrainingvptlearning}:}} We train the inverse dynamics model (IDM) to estimate the relative pose between two consecutive observations such as $p^{i}_{i+1} = f_{idm}(O_i, O_{i+1})$ by imitating the ground-truth relative pose in the dataset. In training, we sample 9 flames from the current frame to the 8-step future frame and estimate the relative poses between each frame by IDM. Then we integrate the estimated relative poses to have the robot trajectories on the current coordinate. 

\noindent
\textit{{\bf GCP~\cite{shah2023vint}:}} 
Following \cite{shah2023vint}, we train the policy as GCP to estimate the robot trajectory to link between two frames, $O_c$ and $O_g$. Since we want to annotate the actions for 8 steps, we select $O_g$ as the 8 step future frame from $O_c$ in training. The others are same as the original paper~\cite{shah2023vint}. 

\setlength{\leftskip}{0.0cm}
\vspace{2mm}
For VPT and GCP, we use both the curated GNM dataset and 1 $\%$ \DatasetName{} dataset to be accurate models. We decide the ratio of the \DatasetName{} dataset as 1 $\%$ according to the data ablation study in the evaluation section \ref{sec:shortnav}. By mixing small \DatasetName{} dataset with the clean GNM dataset, VPT and GCP can suppress the negative effect of the noisy \DatasetName{} dataset and can learn the target robot characteristics. Besides, our \MethodName{} model can use full \DatasetName{} dataset in training due to the robust learning architecture of the model-based learning.
\begin{table}[t]
  \vspace{1mm}
  \begin{center}
  \caption{Evaluation of \PolicyName{} on long-horizon pose-conditioned navigation tasks.  ``GS'' and ``COV'' indicate the goal success rate and the coverage rate.}
  \label{tab:eval_gps}    
  \resizebox{0.9\columnwidth}{!}{
  \begin{tabular}{lclcc} \toprule 
    & \multicolumn{2}{c}{FrodoBots-2K Data} & \multicolumn{2}{c}{Score} \\
    \cmidrule(lr){2-3} \cmidrule(lr){4-5}
    Policy & {Usage} & \hspace{2mm}{Relabeler} & GS & COV \\ \midrule
    NoMaD~\cite{sridhar2024nomad} & GNM only & - & 0.333 & 0.471 \\
    & \checkmark & filtered action~\cite{kfsurvey} & 0.286 & 0.429 \\     
    Behavior Cloning & \checkmark & raw action & 0.286 & 0.567 \\   
    & \checkmark & filtered action~\cite{kfsurvey} & 0.286 & 0.624 \\
    & \checkmark & visual SLAM~\cite{dpvo} & 0.286 & 0.486 \\      
    & \checkmark & VPT~\cite{baker2022videopretrainingvptlearning} & 0.095 & 0.314 \\     
    & \checkmark & GCP~\cite{shah2023vint} & 0.619 & 0.757 \\ 
    \PolicyName{} & \checkmark & \MethodName{} & \textbf{0.857} & \textbf{0.924} \\ \bottomrule
  \end{tabular}%
  }
  \end{center}
\end{table}
\begin{table*}[t]
\vspace{1mm}
\centering
\begin{minipage}[t]{0.35\textwidth}
\centering
\captionof{table}{Quantitative analysis with quadruped robot, Go1 and wheeled robot, VizBot for cross-embodiment analysis.}
    \label{tab:cross_emb}
    \resizebox{1.0\columnwidth}{!}{
    \begin{tabular}{lccccc} 
      \toprule 
      \multicolumn{2}{l}{Method} & \multicolumn{2}{c}{Go1 (outside)} & \multicolumn{2}{c}{VizBot (inside)} \\ 
      \cmidrule(lr){1-2} \cmidrule(lr){3-4} \cmidrule(lr){5-6}
      Policy & Relabeler & GS & COV & GS & COV \\ 
      \midrule
      Behavior Cloning & GCP & 0.300 & 0.680 & 0.200 & 0.630 \\    
      LogoNav & \MethodName{} & {\bf 0.800} & {\bf 0.850} & {\bf 0.600} & {\bf 0.820} \\
      \bottomrule       
    \end{tabular}
    }
\end{minipage}%
\hspace{0.01\textwidth}
\begin{minipage}[t]{0.32\textwidth}
\centering
\captionof{table}{Evaluation of \MethodName{} on action-free in-the-wild YouTube videos. ``GS'' and ``SC'' indicate the goal success rate and the subgoal coverage rate.}
  \label{tab:wildvideo}  
  \resizebox{1.0\columnwidth}{!}{
  \begin{tabular}{clcc} \toprule 
    GNM & YouTube video (LeLaN) & GS & SC \\ \midrule
    \checkmark & \hspace{3mm} \ding{55} & 0.500 & 0.680 \\
    \checkmark  & \hspace{3mm} \checkmark (Visual SLAM~\cite{dpvo}) & 0.125 & 0.313 \\    
    \checkmark  & \hspace{3mm} \checkmark (MBRA) & 0.875 & 0.909 \\
    
    \bottomrule
  \end{tabular}
  }
\end{minipage}%
\hspace{0.01\textwidth}
\begin{minipage}[t]{0.3\textwidth}
\centering
\captionof{table}{Evaluation of the goal image-conditioned navigation at six countries.}
  \label{tab:eval_world} 
  \resizebox{1.0\columnwidth}{!}{
  \begin{tabular}{llcc} \toprule 
    Policy & Dataset & GS & SC\\ \midrule
    GCP~\cite{shah2023vint} & GNM & 0.500 & 0.736 \\
    GCP~\cite{shah2023vint} & GNM + \DatasetName{} (1$\%$) & 0.792 & 0.906 \\ \midrule    
    \MethodName{} & GNM & 0.833 & 0.899 \\
    \MethodName{} & GNM + \DatasetName{} (full) & \textbf{0.958} & \textbf{0.983} \\\bottomrule
  \end{tabular}%
  }
\end{minipage}
\end{table*}

\subsection{Long-horizon Navigation Policy (LogoNav): GPS Goals}
To answer {\bf Q1}, we evaluate the long-horizon navigation policies trained with \MethodName{} and several baselines. We select 7 outdoor locations and evaluate each policy 3 times for each goal. In Table~\ref{tab:eval_gps}, we show the goal success rate and the coverage rate for each method. The coverage rate is the ratio of the distance reached by the robot to the distance of the target goal pose before it fails. Our policy with \MethodName{} shows stronger performance than all baselines for both goal success rate and coverage rate. In Sec.~\ref{sec:shortnav}, we conduct an investigation to analyze the advantageous gap of \MethodName{} to answer {\bf Q3}.

\begin{figure*}[t]
\centering
\includegraphics[width=0.99\hsize, height=0.14\hsize]{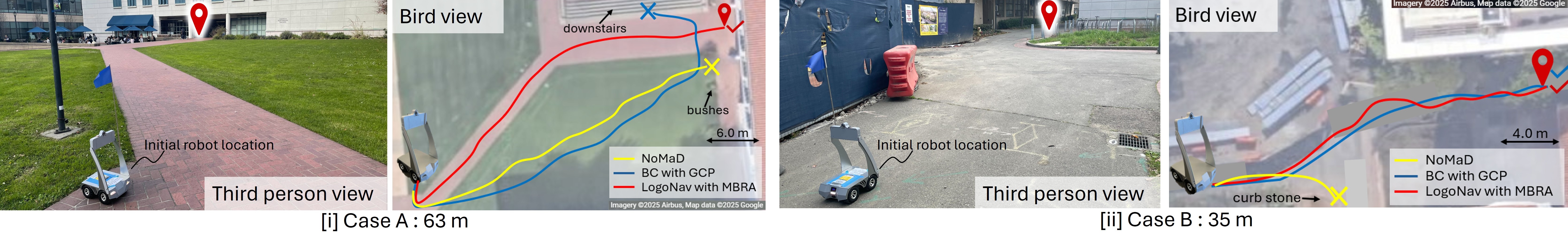}
\caption{{\small\textbf{Policy rollouts for goal pose-conditioned navigation with long-horizon policies.} Our policy, \PolicyName{} trained with \MethodName{} can keep traveling on the road and arrive at the goal pose.}}
\label{f:vis_gps}
\end{figure*}

\begin{figure*}[t]
\centering
\includegraphics[width=0.99\hsize, height=0.19\hsize]{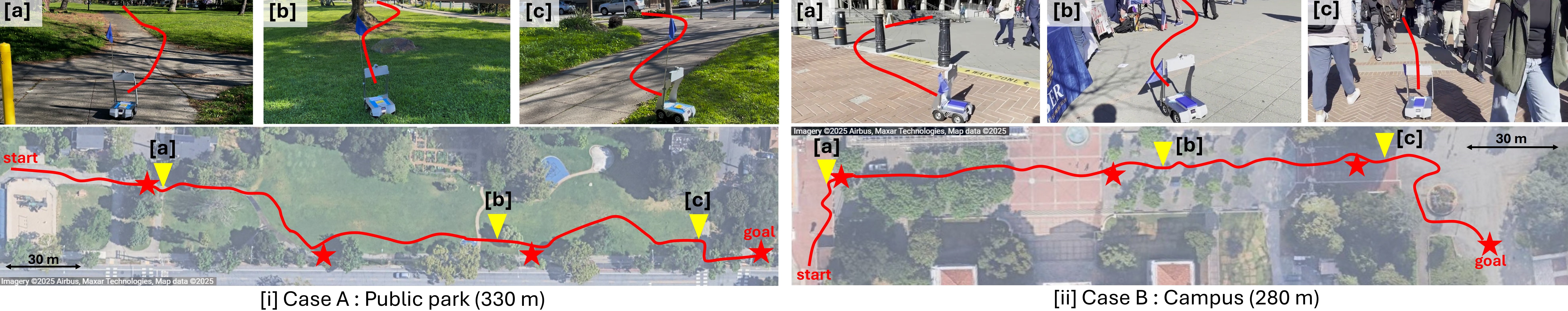}
\caption{{\small\textbf{Long-horizon navigation with multiple subgoals.} The ERZ can travel for about 20 minutes without collision and arrive at the goal about 300 m away. The red stars indicate the subgoal locations.}}
\label{f:long_dist}
\end{figure*}

Figure~\ref{f:vis_gps} shows the third-person view at the start position and the robot trajectories on a bird-eye-view map in two scenes. Our policy distilled from \MethodName{} actions was the only one to successfully navigate to the distant goal pose in both scenes, making a sharp left turn at the start to stay on path in case A. In contrast, both NoMaD and GCP could not execute this action, failing by colliding with bushes or requiring interventions to avoid falling down stairs.
To show the capability of \MethodName{}, we provide several subgoals, specified by latitude, longitude, and azimuth angle values, at intervals of approximately 80 meters, and evaluate LogoNav with \MethodName{} on traversing these subgoals in two scenes. As shown in Fig.~\ref{f:long_dist}, our navigation system with our policy enables us to navigate the robot toward a goal 300 meters away without collision, even in human-occupied spaces.

\begin{figure}[t]
  \centering
  \includegraphics[width=0.99\hsize, height=0.2\hsize]{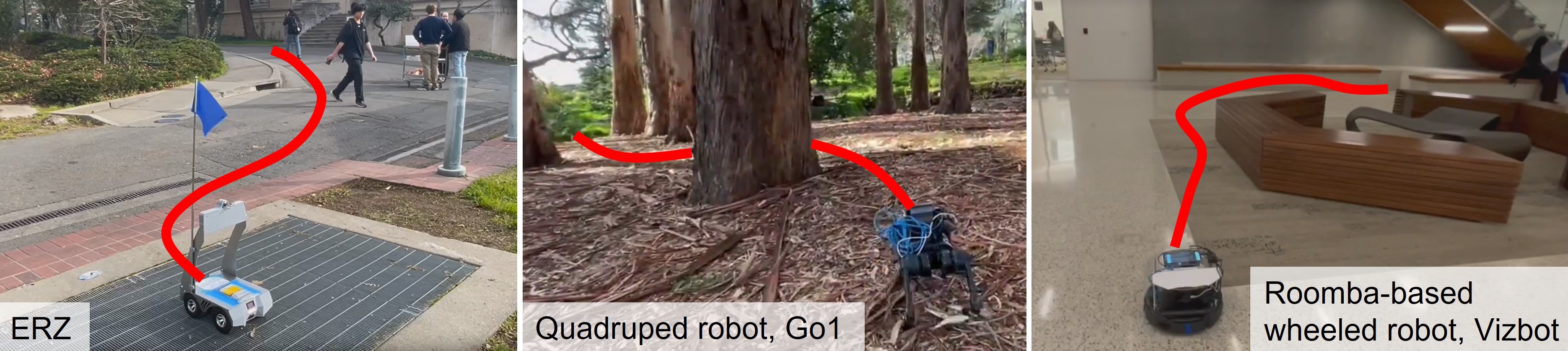}
  \captionof{figure}{{\small\textbf{Visualization of cross-embodiment analysis.}}}
  \label{f:cross_emb}
\end{figure}

Moreover, we deployed LogoNav on two more robotic embodiments, including VizBot~\cite{niwa2022spatio} in an indoor setting, and the Unitree Go1 quadruped robot in an outdoor setting. We conducted 10 trials from up to 100 meters away in different challenging environments with some obstacles for each embodiment and method. We show the quantitative results in Table~\ref{tab:cross_emb} and show the robotic behaviors in Fig.~\ref{f:cross_emb} and the supplemental videos. We achieve strong goal-reaching behavior with collision avoidance compared to the strongest baseline in Table~\ref{tab:eval_gps}, highlighting the policy's generalization ability. Note that we apply the same policy in Table~\ref{tab:eval_gps} and feed the generated actions without any adaptation. Action conversion is internally applied in each robot setup. 

\subsection{Training policies on in-the-wild video with \MethodName{}}

For {\bf Q2}, we evaluate the capability of \MethodName{} on different passive data sources, action-free in-the-wild video. We use the \MethodName{} model to generate the action labels for the in-the-wild videos and train the short-horizon visual navigation policy conditioned on goal images, $\{ a^{v}_i \}_{i=0 \ldots N-1} = \pi^{v}(O_c, O_g)$. During training, we use the same objective $J_{il}$ to imitate the action labels generated by \MethodName{}. We train three goal image-conditioned policies, one with the GNM dataset alone and another two with GNM + in-the-wild videos with different annotations, visual SLAM~\cite{dpvo} and our \MethodName{} model to evaluate how well \MethodName{} enables us to close the embodiment gap between robot and in-the-wild data.

To evaluate the performance in a variety of situations, we collect the topological memories on four indoor trajectories and four outdoor trajectories and deploy the policies with the ERZ. The distance from the initial node to the goal node is between 10.0 m and 31.0 m. As shown in Table~\ref{tab:wildvideo}, the policy trained with the \MethodName{}-annotated in-the-wild video data has an explicit advantage compared to the policy trained only on the GNM dataset. Although the training dataset does not contain the data from the target ERZ, we achieve a high success rate by training with diverse video data.  Besides, visual SLAM often fails on videos with fisheye lenses, dense crowds, or few features, leading to worse performance than the others.
\subsection{Evaluating \MethodName{} for effective crowd-sourced data use}
\label{sec:shortnav}
To answer {\bf Q3}, we compare MBRA and GCP that demonstrated the strongest performance in Table \ref{tab:eval_gps}. We train several relabelers with different data setups for each method and deploy them as the short-horizon navigation policy in the same eight environments and topological memories as in the previous section to more thoroughly explore the capabilities of each of these relabelers.

\begin{table}[t]
  \caption{Comparison of \MethodName{} and GCP on short-horizon navigation.}
  \begin{center}
  \label{tab:comparison_act}    
  \resizebox{0.85\columnwidth}{!}{
  \begin{tabular}{clcccc} \toprule 
    \multicolumn{2}{c}{Dataset} & \multicolumn{2}{c}{GCP} & \multicolumn{2}{c}{\MethodName{}} \\ 
    \cmidrule(lr){1-2} \cmidrule(lr){3-4} \cmidrule(lr){5-6}
    GNM & \DatasetName{} & GS & SC & GS & SC \\ \midrule
    \checkmark & \ding{55}& 0.500 & 0.680 & 0.875 & 0.960 \\    
    \checkmark & raw action & 0.000 & 0.308 & 0.500 & 0.777 \\
    \ding{55}  & filtered action & 0.125 & 0.377 & 0.875 & 0.940 \\
    \checkmark & filtered action(1$\%$) & 0.750 & 0.887 & 0.875 & 0.889 \\   
    \checkmark & filtered action(10$\%$) & 0.375 & 0.638 & 0.875 & 0.970 \\
    \checkmark & filtered action(40$\%$) & 0.500 & 0.641 & \textbf{1.000} & \textbf{1.000} \\
    \checkmark & filtered action(70$\%$) & 0.375 & 0.748 & \textbf{1.000} & \textbf{1.000} \\    
    \checkmark & filtered action & 0.375 & 0.576 & \textbf{1.000} & \textbf{1.000} \\
    \bottomrule       
  \end{tabular}%
  }
  \end{center}
\end{table}
Table~\ref{tab:comparison_act} shows the goal success rate and the subgoal coverage rate for each policy. We find that GCP completely deteriorates the performance by imitating the noisy raw action of \DatasetName{} dataset. The EKF filtering helps a bit, and incorporating the GNM data improves performance as well. In our data ablation study, we find that GNM + only 1$\%$ \DatasetName{} dataset can help to improve the performance. However, GCP cannot effectively leverage the entire \DatasetName{} dataset. Besides, MBL enables us to scalably learn our \MethodName{} model from the noisy data. \MethodName{} model trained on GNM + filtered $100\%$ \DatasetName{} dataset successfully arrived at the goal position in all cases. 

In the final experiment, we aim to assess the generalization capabilities of \MethodName{} model. To this end, we deploy the short-horizon navigation policy on robots in diverse environments across 6 countries: USA, Mexico, China, Mauritius, Costa Rica, and Brazil. In total, we collect 24 topological graphs and evaluate each trajectory. To the best of our knowledge, we are the first to conduct a global evaluation for visual navigation. We evaluate GCP and \MethodName{} model trained with and without the \DatasetName{} dataset. Findings are summarized in Table \ref{tab:eval_world}. \MethodName{} model as short-horizon goal image-conditioned navigation policy had better performance for both goal reaching and subgoal coverage than GCP. 

\section{Conclusion} 
\label{sec:conclusion}
\MethodName{} allows us to leverage large amounts of low-quality passive data for learning long-horizon navigation policies, making affordable passive data useful for training broadly generalizable and capable visual navigation policies. \MethodName{} trains a short-horizon image-conditioned navigation policy to reannotate imprecise trajectory action labels. Then, the reannotated labels are used as ground truth to train a goal-pose conditioned long-horizon policy, which learns reasonable conventions such as staying on paths and avoiding collisions. We evaluate our method on robots in 6 countries across multiple continents and observe significant improvements over baselines. These results indicate that our model provides a broadly applicable, capable, and generalizable solution for visual navigation.

\smallskip \noindent \textbf{Limitations:} Our \MethodName{} approach for reannotating noisy crowd-sourced data and action-free in-the-wild videos for long-horizon navigation conditioned on a 2D goal pose works well, and we confirmed its applicability to other navigation tasks~\cite{hirose2025omnivla}, though it still leaves room for improvement.
In the model-based approach, we may sometimes generate unreasonable actions because of inaccuracies in the robot model. While we find the model-based approach to generally outperform the imitation-based relabeler (GCP), it does require some strong conditions on the model itself that could prove difficult to translate to more complex tasks like manipulation. A promising direction for future work is developing a more accurate differentiable model that incorporates richer 3D geometry, semantic information, and dynamic object behaviors (e.g., pedestrian motion~\cite{hirose2023sacson}), which are essential for robust navigation in cluttered indoor environments. Another important extension is to account not only for goal reaching but also for human preferences—particularly in crowded spaces or settings with strong semantic norms (e.g., avoiding grass when inappropriate). While our model inherits some of these behaviors from human operators in the data, such preferences are not explicitly enforced.

\bibliographystyle{IEEEtran}
\vskip-\parskip
\begingroup
\footnotesize
\balance
\bibliography{references_detail}
\endgroup
\vfill

\end{document}